\documentclass[10pt,letterpaper]{article}

\usepackage{cogsci}
\usepackage{graphicx}
\usepackage{booktabs}
 \usepackage{multirow}
 \usepackage{algorithm}
\usepackage{algpseudocode}  % from algorithmicx
\cogscifinalcopy %%% Uncomment this line for the final submission

\usepackage[
  style=apa,
  natbib=true,
  annotation=false,
]{biblatex}
\usepackage{float} %%% Roger Levy added this and changed figure/table placement to [H] for conformity to Word template, though floating tables and figures to top is still generally recommended!

\usepackage{todonotes}

\title{Fine-Refine: Iterative Fine-grained Refinement for Mitigating Dialogue Hallucination}

\author[1]{\mbox{Xiangyan Chen (xiangyan.chen@qmul.ac.uk)}}
\author[2]{\mbox{Yujian Gan (y.gan@qub.ac.uk) }}
\author[1,3]{\mbox{Matthew Purver (m.purver@qmul.ac.uk)}}
\affil[1]{Queen Mary University of London, UK}
\affil[2]{Queen's University Belfast, UK}
\affil[3]{Institut Jožef Stefan, Slovenia}

\begin{document}

\maketitle

\begin{abstract}
The tendency for hallucination in current large language models (LLMs) negatively impacts dialogue systems. Such hallucinations produce factually incorrect responses that may mislead users and undermine system trust. Existing refinement methods for dialogue systems typically operate at the response level, overlooking the fact that a single response may contain multiple verifiable or unverifiable facts. To address this gap, we propose \textsc{Fine-Refine}, a fine-grained refinement framework that decomposes responses into atomic units, verifies each unit using external knowledge, assesses fluency via perplexity, and iteratively corrects granular errors. We evaluate factuality across the HybriDialogue and OpendialKG datasets in terms of factual accuracy (fact score) and coverage (Not Enough Information Proportion), and experiments show that \textsc{Fine-Refine} substantially improves factuality, achieving up to a 7.63-point gain in dialogue fact score, with a small trade-off in dialogue quality.%\todo{"with a small trade-off"??} 

%We evaluate factuality using dialogue fact score and Not Enough Information Proportion (NEIP), and assess dialogue quality with coherence and fluency judgments. We evaluate factuality across the HybriDialogue and OpendialKG datasets using the dialogue fact score, and the results show that our proposed method markedly improves factual accuracy, achieving up to a 7.63-point increase in the dialogue fact score.

\end{abstract}

\section{Introduction}
\label{sec:intro}
Large language models (LLMs) are a core component of modern dialogue systems. However, they can generate hallucinated responses that are fluent and plausible yet factually incorrect or unsupported by evidence. Such outputs may mislead users by creating an illusion of reliability, raising concerns about trust and safe deployment.

Retrieval-augmented generation (RAG) \citep{baek2023knowledge, he2024g, chen2025improving, du2025evidence} has mitigated hallucinations in knowledge-grounded dialogue and open-domain question answering (QA) by conditioning generation on external evidence. However, most methods lack explicit post-generation verification and refinement steps, and therefore cannot guarantee factual correctness. Existing refinement approaches, such as \textsc{Self-Refine} \citep{madaan2023self}, aim to improve the quality of dialogue responses by iterative refinement. But refinement operates at the coarse response level, where a single response may contain both correct and incorrect facts, making feedback noisy and inefficient.

\begin{figure*}[t]
\center
\includegraphics[width=1\linewidth]{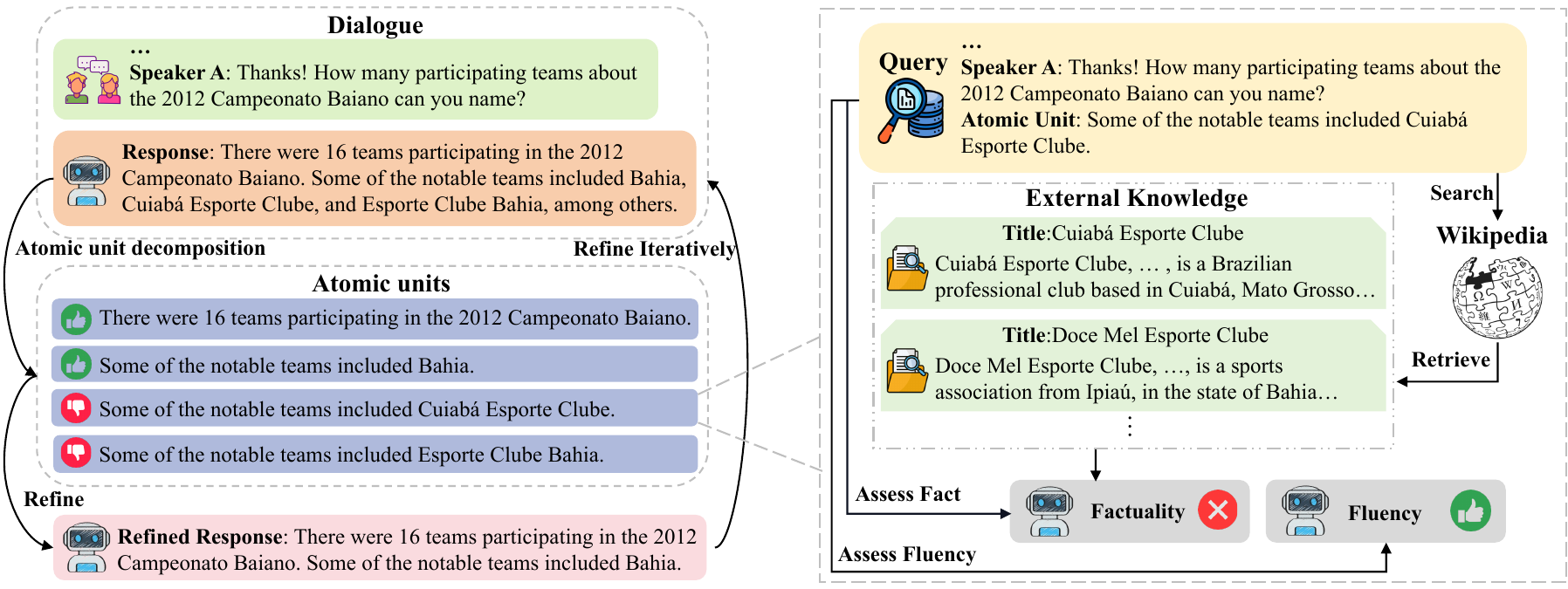}
\caption{Overall framework of the proposed fine-grained refinement. Starting from the dialogue, the original LLM-generated response contains several factually incorrect atomic units. The right part describes the assessment details. After assessing these atomic units, the LLM refines the response accordingly.}
%\caption{Fine-grained refinement decomposes a response into atomic units, enabling iterative feedback and improvement.}
\label{fig:example}
\end{figure*}

To address the gap, we propose a fine-grained, iterative refinement framework, named \textsc {Fine-Refine}, inspired by the subgoaling strategy \citep{simon1971human}, by which humans solve problems by decomposing complex questions into smaller tasks.%\todo{Maybe try to make it very clear this is about human minds: e.g. "by which humans solve problems by decomposing [etc]"?} 
Given a generated response, we decompose it into atomic units via a %\todo{"via a LLM"} 
LLM. We define an atomic unit as the smallest unit of information that cannot be decomposed further. Figure~\ref{fig:example}(left) shows a dialogue about \emph{2012
Campeonato Baiano}, a response and its decomposed atomic units. The atomic units are assessed for quality based on factuality and fluency. For the factuality assessment, each atomic unit is verified via external knowledge by LLM, with the external knowledge sourced from Wikipedia. Fluency assessment is based on the measurement of perplexity \citep{jelinek1977perplexity}. We use the assessment results as feedback to refine the response with the LLM. Since a single refinement pass may not fully resolve all factual errors, we perform iterative refinement guided by this feedback. To ensure efficiency, we cap the process at N iterations, yielding the final refined response.

We evaluate our framework on the HybriDialogue \citep{nakamura2022hybridialogue} and OpendialKG \citep{moon2019opendialkg} datasets, comparing against RAG, refinement, and LLM baselines using automatic metrics for factuality, and LLM-based judgements of coherence and fluency. Since there is no unified LLM judge for automatic coherence and fluency evaluation, we benchmark popular open-source LLMs under the G-Eval \citep{liu-etal-2023-g} setting against the human annotations released by \citet{mehri-eskenazi-2020-usr} on PersonaChat \citep{zhang2018personalizing} and Topical-Chat \citep{gopalakrishnan2023topical}. We then select the judge that best aligns with humans for all subsequent evaluations, enabling reproducible evaluation with publicly available models.

We use the dialogue fact score and the Not Enough Information Proportion (NEIP) to assess factuality, as validated in prior work \citep{chen2025improving}. Across HybriDialogue and OpendialKG datasets, \textsc{Fine-Refine} consistently improves factual accuracy, with up to a 7.63-point gain in dialogue fact score on HybridDialogue with Qwen3-8B \citep{yang2025qwen3}, with a small trade-off in dialogue quality. It indicates effective hallucination mitigation by reducing fact-inconsistent generations in multi-turn dialogue.

Our contributions are as follows:

\begin{enumerate}

\item We propose a novel, fine-grained, iterative assessment and refinement framework inspired by the subgoaling strategy from cognitive psychology. %\todo{again maybe "inspired by the subgoaling strategy from human psychology", or "used by humans" or similar?} 
Our framework primarily aims to mitigate hallucinations in dialogue response generation by providing fine-grained feedback to guide the LLM’s refinement.

\item We benchmark several open-source LLMs under the G-Eval protocol on PersonaChat and Topical-Chat against human annotations, and select the best-aligned model as our evaluator for dialogue quality.

\item Experiments on HybriDialogue and OpendialKG datasets demonstrate that our framework markedly improves factuality with various LLMs, outperforming baselines by up to 7.63 points and enhancing reliability in knowledge-grounded dialogue generation.

\end{enumerate} 

\section{Related Work}
Most recent works mainly focus on RAG, which uses retrieved knowledge to generate text that effectively mitigates hallucination. In addition, some works used post-editing methods. 

\subsection{Retrieval-augmented Generation}
\citet{baek2023knowledge} aimed to answer this question using a knowledge graph. They adopt an arbitrary pre-trained language model to reason the relationship between the question and retrieved knowledge triples. The retrieved knowledge is used for improving the factuality of generation. Self-RAG \citep{asai2024self} trains reflection tokens, including relevance, usefulness, and support, based on GPT-4o-generated data, which helps determine when to retrieve and whether the question and knowledge are relevant. In addition, \citet{he2024g} proposed a graph-based encoder that connects the retrieved subgraph and the question, helping retrieve the most relevant knowledge. \citet{du2025evidence} explored possible combinations of question, evidence, and answer, which help understand the logical connections among them and further mitigate hallucinations. In dialogue systems, \citet{chen2025improving} proposed a framework that integrates coreference resolution, dialogue-sense knowledge selection, and a graph-based encoder to effectively encode dialogue and knowledge, thereby improving dialogue factuality.

However, these works mainly focus on the knowledge-grounded generation, which lacks the post-generation verification and refinement. 

\subsection{Refinement-based Approaches}
Self-refine \citep{madaan2023self} is an iterative self-correction framework that generates a response, then verifies it against multiple quality dimensions, such as coherence and fluency, and refines it based on the verification results. It shows an impressive improvement in diverse tasks, including dialogue response generation and mathematical reasoning. However, \citet{huanglarge} declares that the model cannot be corrected by itself without external knowledge, implying the importance of external knowledge. To mitigate hallucination, \citet{lee2024ask} proposed an \(A^2R\) framework that asks, assesses, and refines the generated text with metric-guided feedback, thereby improving factuality in abstract summarisation. 

However, these works mainly focus on the response level. But relying on the response level to verify the fact is difficult, especially for long-form dialogue responses. \citet{simon1971human} discussed the subgoaling strategy in human problem-solving,%\todo{"in human problem-solving"?} 
which splits a complex problem into smaller tasks. Inspired by this, our framework addresses the gap by introducing fine-grained refinement. 

\section{Methodology}
\label{sec:format}
Previous works, such as RAG \citep{he2024g, asai2024self}, lack post-generation verification. Post-generation methods like \textsc{Self-Refine} focus only on response-level refinement, which is often insufficient for improving factual accuracy, as a single response may contain multiple verified or unverified facts. 

To address this gap, we propose a novel fine-grained and iterative refinement method, \textsc{Fine-Refine}, for dialogue responses. Our approach leverages LLMs to generate fine-grained feedback and iteratively refines the generated response based on this feedback. Figure~\ref{fig:example} provides an overview of the framework: the LLM’s initial response contains factual errors, which we correct through iterative refinement. Specifically, we first decompose the response into atomic units and then assess each unit individually—factuality is verified against external knowledge using an LLM, while fluency is measured by perplexity. These atomic units, together with their assessment results, constitute fine-grained feedback that guides the LLM to iteratively refine the response.

\subsection{Problem Definition}
Given a dialogue \(\mathcal{U}=(U_1, U_2, ..., U_t)\), where \(U_t\) is the initial generated response from the LLM \(\mathcal{M}_{\theta}\), we aim to refine \(U_{t}\) into refined response \(U^{'}_{t}\) given the \(\mathcal{U}\) and \(U_t\). In our refinement method, we decompose the response into several atomic units \(\mathcal{T}=(t_1, t_2, ..., t_n)\), where \(n\) denotes the number of atomic units. We define an atomic unit as a factual proposition in a response. For each atomic unit \(t_i\), we retrieve \(K\) passages \(\mathcal{P}_i=\{p_1, p_2, ..., p_K\}\), and verify the corresponding factuality \(\mathcal{F}=(f_1, f_2, ..., f_n)\) for atomic units \(\mathcal{T}\), where \(f_i \in \{\emph{Supports}, \emph{Refutes}, \emph{Not Enough Information}\}\). We also compute the fluency scores for \(\mathcal{T}\), denoting \(\mathcal{E}=(e_1, e_2, ..., e_n)\).

\subsection{Atomic Unit Decomposition}
\label{sec:atomic_fact_decomposition}
A long-form dialogue response may contain multiple factual claims, even within a single sentence, which motivates a more fine-grained claim-level decomposition.

We decompose the dialogue response into several pieces --- atomic units --- by an arbitrary LLM \(\mathcal{M}\) with the trained parameters \( \theta \). To ensure the decomposition performance, few-shot learning is applied. Specifically, we select two demonstrations based on the BM25 \citep{robertson2009probabilistic} search between the input response and the hand-crafted demonstration database. With a manually designed prompt \(r_a\) that decomposes the response \(U_t\) into atomic units \(\mathcal{T}\), the process can be formulated as:

\begin{equation}
    \mathcal{T}=\mathcal{M}_{\theta}(r_{a}, U_t).
\end{equation}

\subsection{Factuality Assessment}
\label{sec:factuality_assessment}
We verify each atomic unit using an LLM with Chain-of-Thought (CoT) prompting \citep{wei2022chain} and retrieved external evidence. Since we verify dialogue factuality, the dialogue context is included in this work. 

For each atomic unit, we retrieve the corresponding passages as external knowledge from the Contriever-MS MARCO \citep{izacard2021unsupervised}, denoted as \(\mathcal{R}\), a dense retriever designed on contrastive learning and showing a strong performance on retrieval. We retrieve the passages \(\mathcal{P}_i\) from Wikipedia dump data based on the concatenation of dialogue \(\mathcal{U}\) and the atomic unit \(t_i\). The retrieval process with top-$K$ selection is defined as:
\begin{equation}
    \mathcal{P}_i = \operatorname{TopK}\big(\mathcal{R}(\mathcal{U}, t_i), K\big).
\end{equation}

We verify each \(t_i\) via \(\mathcal{M}_{\theta}\), given the passages \(\mathcal{P}_i\), \(\mathcal{U}\) and \(t_i\). \(\mathcal{M}_{\theta}\) is the same LLM \(\mathcal{M}_{\theta}\) used in the aforementioned atomic unit decomposition. Following CoT prompting, we encourage the LLM to perform step-by-step reasoning before producing the final verification label. Such intermediate reasoning has been widely studied and is often associated with improved performance on reasoning-intensive tasks, including fact verification. The reasoning-based fact verification is formulated as follows:

\begin{equation}
    f_i=\mathcal{M}_{\theta}(r_f,\mathcal{U}, t_i, \mathcal{P}_i),
\end{equation}
where \(r_{f}\) is the fact verification prompt and \(f_i \in \{\emph{Supports}, \emph{Refutes}, \emph{Not Enough Information}\}\) is the factual result. ``Supports'' denotes the atomic unit that is supported by retrieving the external knowledge and dialogue context, while ``Refutes'' denotes the opposite, i.e., the evidence contradicts the atomic unit. ``Not Enough Information'' means the atomic unit cannot find the external knowledge to verify, or the retrieved knowledge is insufficient to verify. Following prior work on CoT prompting \citep{wei2022chain}, we add an instruction such as ``think step by step'' during verification. We only retain the model’s final label as output, as including full CoT rationales would substantially increase feedback length and introduce noisy intermediate statements, which complicates iterative refinement prompts while providing limited additional benefit for targeted edits.

\subsection{Fluency Assessment}
\label{sec:fluency_assessment}
The dialogue quality assessment is also included to prevent fact-driven edits that degrade readability. We consider the fluency score in this work.

We compute the perplexity \(\mathcal{Q}=(q_1, q_2, ..., q_n)\) for atomic units \(\mathcal{T}\) by GPT-2 \citep{radford2019language},  where lower perplexity generally indicates higher fluency. We then map the perplexity value \(q_i\) to a normalised fluency score \(e_i\) as follows:

\begin{equation}
    e_i=1-\frac{q_{i}-min(\mathcal{Q})}{max(\mathcal{Q})-min(\mathcal{Q})+\epsilon}.
\end{equation}

Here, $\min(\cdot)$ and $\max(\cdot)$ denote the minimum and maximum operators, respectively. This normalisation yields a relative fluency score within the set of atomic units, facilitating comparison across units in the same response. The small constant $\epsilon$ is introduced to avoid division by zero.

\subsection{Iterative Feedback-Guided Response Refinement}
\label{sec:iterative_feedback_guided_response_refinement}
Fine-grained feedback is the core of our proposed framework, which is formulated as a structured critique that explicitly links each atomic unit in the response to its quality assessment. The combination of fine-grained feedback and dialogue context guides the LLM to refine its response. However, a single-time refinement may be insufficient to correct all incorrect atomic units or to prevent the introduction of new partial hallucinations. To solve the limitations of the single-time approach, we refine the response iteratively.

Fine-grained feedback includes atomic units \(\mathcal{T}\), and their assessment , such as the factual label \(\mathcal{F}\) and fluency score \(\mathcal{E}\), which is defined as:

\begin{equation}
    S=\mathcal{T} \oplus \mathcal{F} \oplus \mathcal{E},
\end{equation}
where \(\oplus\) aggregates the components into a single feedback string \(S\), which is included in the prompt context for refinement.

We use the same LLM \(\mathcal{M}_\theta\) to perform refinement via prompting, and define the iterative refinement process as follows:
\begin{equation}
    U^{(1)}_{t}=\mathcal{M}_{\theta}(r_g, \mathcal{U}, U_t, S),
\end{equation}

\begin{equation}
U^{(l+1)}_t=\mathcal{M_\theta}(r_g, \mathcal{U},U^{(l)}_t,S^{(l)}),
\end{equation}

\begin{equation}
U'_t=U_t^{(L)}.
\end{equation}

Here, \(r_g\) is the prompt for refinement, which asks the LLM to refine the original response based on the fine-grained feedback \(S\). The subscript of \((l)\) denotes the refinement iteration. \(S^{(l)}\) is the feedback derived from \(U^{(l)}_t\). Each iteration recomputes \(\mathcal{T}\),\(\mathcal{F}\), and \(\mathcal{E}\) based on the updated response. We refine the response when the iteration number reaches the maximum iteration count \(L\) and get the final response \(U^{'}_{t}\).

\setlength{\tabcolsep}{1.5pt}
\begin{table}[t]
\caption{We report correlations (r: Pearson, $\rho$: Spearman) on PersonaChat and Topical-Chat. \emph{Humans} denotes the average of three human–human correlations, and model scores are averaged against humans. \emph{Maintains Ctx.} denotes Maintains Context. Best results are in bold. }
\centering
\resizebox{0.49\textwidth}{!}{
\begin{tabular}{l|cc|cc|cc|cc}
\toprule
\multirow{3}{*}{\textbf{Methods}} 
& \multicolumn{4}{c|}{\textbf{PersonaChat}} 
& \multicolumn{4}{c}{\textbf{Topical-Chat}} \\
\cmidrule(lr){2-5} \cmidrule(lr){6-9}
& \multicolumn{2}{c|}{\textbf{Maintains Ctx.}} & \multicolumn{2}{c|}{\textbf{Natural}} 
& \multicolumn{2}{c|}{\textbf{Maintains Ctx.}} & \multicolumn{2}{c}{\textbf{Natural}} \\
\cmidrule(lr){2-3} \cmidrule(lr){4-5} \cmidrule(lr){6-7} \cmidrule(lr){8-9}
& \textbf{r} & \textbf{$\rho$} & \textbf{r} & \textbf{$\rho$} & \textbf{r} & \textbf{$\rho$} & \textbf{r} & \textbf{$\rho$} \\
\midrule
Humans & 0.613 & 0.612 & 0.472 & 0.484 & 0.557 & 0.560 & 0.486 & 0.487 \\
\midrule
Llama-3.3-70B & 0.521 & 0.561 & 0.297 & 0.317 & 0.457 & \textbf{0.512} & 0.447 & 0.453 \\
Qwen3-32B & 0.529 & 0.509 & 0.325 & 0.325 & \textbf{0.486} & 0.494 & 0.450 & \textbf{0.484} \\
Gemma3-27B & \textbf{0.552} & \textbf{0.556} & \textbf{0.333} & \textbf{0.370} & 0.481 & 0.492 & \textbf{0.465} & 0.466 \\
\bottomrule
\end{tabular}
}

\label{table:chat-results}
\end{table}

\begin{table*}[h]
\centering
\caption{Performance across HybriDialogue and OpendialKG. \emph{Fact} denotes the dialogue fact score. Bold indicates the best per category. Superscripts indicate the number of refinement iterations.}
\resizebox{0.85\textwidth}{!}{
\begin{tabular}{l|cccc|cccc}
\toprule
\multirow{2}{*}{\textbf{Methods}} & \multicolumn{4}{c|}{\textbf{HybriDialogue}} & \multicolumn{4}{c}{\textbf{OpendialKG}} \\
\cmidrule(lr){2-5} \cmidrule(lr){6-9}
& \textbf{Fact} & \textbf{NEIP$\downarrow$} & \textbf{Maintains Ctx.}  & \textbf{Natural} 
& \textbf{Fact} & \textbf{NEIP$\downarrow$} & \textbf{Maintains Ctx.} & \textbf{Natural} \\
\toprule
%CDT \citep{yang2025improving} & 67.89 & 65.04 & 2.721 & 2.706 & 83.58 & 62.22 & 2.801 & 2.851 \\ 
\textbf{G-Retriever }  \citep{he2024g}
& 70.29 & 55.77 & \textbf{2.824} & \textbf{2.788}
& 84.79 & 62.65 & \textbf{2.895} & \textbf{2.927} \\ 
\textbf{Self-RAG} \citep{asai2024self} & \textbf{79.74} & \textbf{46.39} & 2.162 & 2.238 & \textbf{84.06} & \textbf{49.53} & 2.462 & 2.418 \\
\midrule

\textbf{Mistral-7B-Instruct-v0.3 } \citep{jiang2024mistral}
& 77.41 & 50.28 & \textbf{2.935} & \textbf{2.948} 
& 86.79 & 49.60 & \textbf{2.988} & \textbf{2.914} \\ 

+ \textsc{Self-Refine}\textsuperscript{3} \citep{madaan2023self}
& 79.65 & 63.10 & 2.894 & 2.755
& 77.12 & 59.20 & 2.893 & 2.652 \\ 

+ \textsc{Fine-Refine}\textsuperscript{3} (Ours) 
& \textbf{80.62} & \textbf{45.97} & 2.879 & 2.775 
& \textbf{86.81} & \textbf{46.52} & 2.945 & 2.627 \\ 

\midrule

\textbf{Llama-3.1-8B-Instruct}  \citep{dubey2024llama}
& 79.74 & 50.56 & 2.807 & \textbf{2.724} 
& 85.00 & 46.27 & \textbf{2.919} & \textbf{2.690} \\ 

+ \textsc{Self-Refine }\textsuperscript{3} \citep{madaan2023self}
& 81.73 & 61.18 & \textbf{2.883} & 2.598 
& 88.83 & 48.91 & 2.915 & 2.545 \\ 

+ \textsc{Fine-Refine}\textsuperscript{3} (Ours)
& \textbf{85.89} & \textbf{41.62} & 2.734 & 2.552 
& \textbf{88.84} & \textbf{36.51} & 2.859 & 2.512 \\ 

\midrule

\textbf{Qwen3-8B }  \citep{yang2025qwen3}
& 73.22 & 54.26 & \textbf{2.878} & \textbf{2.933}
& 84.16 & 59.36 & \textbf{2.973} & \textbf{2.965} \\ 

+ \textsc{Self-Refine \textsuperscript{3} }\citep{madaan2023self}
& 75.66 & 61.99 & 2.875 & 2.868
& 85.14 & 60.65 & 2.952 & 2.902 \\ 

+ \textsc{Fine-Refine}\textsuperscript{3} (Ours)
& \textbf{80.85} & \textbf{48.37} & 2.810 & 2.835 
& \textbf{87.91} & \textbf{56.99} & 2.962 & 2.949 \\ 

\bottomrule
\end{tabular}
}

\label{table:main-results}
\end{table*}

\section{Experiment}
\label{sec:prior}
\subsection{Dataset}
In this work, we employ the HybriDialogue \citep{nakamura-etal-2022-hybridialogue} and OpendialKG \citep{moon2019opendialkg} datasets, both of which are open-domain, knowledge-grounded dialogue datasets. HybriDialogue covers diverse topics, such as historical events and biology. OpendialKG includes book and movie recommendations. We follow the turn-level evaluation setting in \citet{chen2025improving}: HybriDialogue has 1,111 test turns (from 243 dialogues), and OpenDialKG has 1,973 test instances.

\subsection{Metrics}
We evaluate factuality using the dialogue fact score and the Not Enough Information Proportion (NEIP) \citep{chen2025improving}, with lower NEIP values preferred. The dialogue fact score is computed as the proportion of supported atomic units among all verifiable atomic units, and has been verified with high Cohen's Kappa \citep{cohen1960coefficient} with human annotators \citep{chen2025improving}. NEIP is the proportion that atomic units cannot be verified. 

As noted in previous studies, such as G-Eval \citep{liu-etal-2023-g}, traditional word-overlap metrics, such as BLEU \citep{papineni2002bleu}, often show low correlation with human judgments when assessing dialogue quality. Following G-Eval, we employ an LLM as a judge. But its original evaluation on closed-source models limits reproducibility.

Following the correlation-based evaluation protocol in USR \citep{mehri-eskenazi-2020-usr}, we report Pearson \citep{pearson1895vii} and Spearman \citep{spearman1961proof} correlations on PersonaChat \citep{zhang2018personalizing} and Topical-Chat \citep{gopalakrishnan2023topical}. PersonaChat and Topical-Chat are dialogue datasets, and we use human quality annotations collected in USR \citep{mehri-eskenazi-2020-usr}. We evaluate three open-source LLMs, including Llama-3.3-70B \citep{dubey2024llama}, Qwen3-32B \citep{yang2025qwen3}, and Gemma3-27B \citep{team2025gemma}. Our evaluation focuses on \emph{maintains context}, which assesses whether a response is a valid continuation of the conversation, and \emph{natural}, which assesses whether it could plausibly be said by a person \citep{mehri-eskenazi-2020-usr}. We use maintains context and natural as proxies for the coherence and fluency aspects mentioned in the introduction.

As shown in Table~\ref{table:chat-results}, Gemma3-27B outperforms the other LLMs on both Pearson and Spearman correlations across PersonaChat and Topical-Chat datasets, with consistency to human evaluators that is comparably close. Based on the overall performance, we adopt Gemma3-27B to evaluate both maintains context and natural in our experiments.

\subsection{Baselines}
Mistral-7B-Instruct-v0.3 \citep{jiang2024mistral}, Qwen3-8B \citep{yang2025qwen3} and Llama-3.1-8B-Instruct \citep{dubey2024llama} are adopted as LLM backbones for all prompting-based baselines. We compare our method with the baseline method \textsc{Self-refine} \citep{madaan2023self}, G-Retriever \citep{he2024g}, Self-RAG \citep{asai2024self}, as listed as follows:

\begin{itemize}\itemsep=0pt

\item Self-Refine generates an initial response with an LLM, then uses the same LLM to provide self-feedback along multiple quality dimensions (e.g., coherence and fluency), and iteratively refines the response based on this feedback.

\item G-Retriever is a graph-based retrieval-augmented generation framework that retrieves a relevant subgraph from a knowledge graph, encodes it with a graph encoder, and conditions an LLM decoder on the encoded subgraph to generate the response. We adopt Llama-2-7B as the backbone model, consistent with prior work.

\item Self-RAG is a text-based retrieval-augmented generation approach that learns reflection/critique signals to decide when to retrieve and to assess whether the generated response is supported by the retrieved evidence. We adopt Llama-2-7B as the backbone, following prior work

\end{itemize}

\subsection{Experimental Setup}
To ensure a fair comparison, we control the decoding configuration (e.g., max tokens and sampling strategy) across methods, and use the same retrieval corpus, retriever, and top-K passages whenever external evidence is required. 

Moreover, since the original \textsc{Self-Refine} feedback does not explicitly account for factuality, we augment it with an evidence-grounded factuality judgment using the same verifier and retrieved passages as in \textsc{Fine-Refine}, while keeping the refinement prompt format and iteration budget unchanged.

Regarding computational cost, our framework introduces additional overhead by decomposing each response into an average of 8.22 atomic units. For each atomic unit, we retrieve the top-4 passages as evidence and perform one LLM-based verification; the aggregated unit-level feedback is then used to refine the response (once per iteration).

\begin{table*}[!ht]
\caption{Qwen3-8B results on HybriDialogue and OpendialKG with refinement iteration counts in superscript.}
\centering
\resizebox{0.76\textwidth}{!}{
\begin{tabular}{l|l|cccc|cccc}
\toprule
\multirow{2}{*}{\textbf{Model}} & \multirow{2}{*}{\textbf{Iteration}} & \multicolumn{4}{c|}{\textbf{HybriDialogue}} & \multicolumn{4}{c}{\textbf{OpendialKG}} \\
\cmidrule(lr){3-6} \cmidrule(lr){7-10}
 & & \textbf{Fact} & \textbf{NEIP$\downarrow$} & \textbf{Maintains Ctx.} & \textbf{Natural}
   & \textbf{Fact} & \textbf{NEIP$\downarrow$} & \textbf{Maintains Ctx.} & \textbf{Natural} \\
\toprule
\

\multirow{3}{*}{Qwen3-8B} 
 & \textsc{Self-Refine}\textsuperscript{1} & 61.31 & 59.70 & 1.036 & 1.038 & 74.47 & 76.48 & 1.252 & 1.331 \\
 & \textsc{Self-Refine}\textsuperscript{2} & 74.70 & \textbf{59.38} & 2.854 & \textbf{2.871} & 84.68 & \textbf{58.99} & 2.925 & \textbf{2.902} \\
 & \textsc{Self-Refine}\textsuperscript{3} & \textbf{75.66} & 61.99 & \textbf{2.875} & 2.868 & \textbf{85.14} & 60.65 & \textbf{2.952} & \textbf{2.902} \\
\midrule
\multirow{3}{*}{Qwen3-8B} 
 & \textsc{Fine-Refine}\textsuperscript{1} & 79.39 & 50.08 & \textbf{2.842} & \textbf{2.881} & 87.36 & 57.62 & \textbf{2.970} & \textbf{2.961} \\
 & \textsc{Fine-Refine}\textsuperscript{2} & 80.74 & 49.43 & 2.820 & 2.844 & 87.55 & 57.15 & 2.965 & 2.953 \\
 & \textsc{Fine-Refine}\textsuperscript{3} & \textbf{80.85} & \textbf{48.37} & 2.810 & 2.835 & \textbf{87.91} & \textbf{56.99} & 2.962 & 2.949 \\
\bottomrule
\end{tabular}
}
\label{table:iterative_models_llm}
\end{table*}

\subsection{Main Results}
The main results of our framework, along with the RAG and refinement baselines, are presented in Table~\ref{table:main-results}.

Regarding factuality, \textsc{Self-Refine} improves the dialogue fact score for all backbone models except Mistral-7B-Instruct-v0.3 on OpendialKG, and it typically increases NEIP (NEIP means the proportion of atomic units with not enough information; lower is better). In contrast, our \textsc{Fine-Refine} continuously improves factuality shown as the increase of dialogue fact score, and decrease of NEIP. The largest dialogue fact score gain is an absolute +7.63 points for Qwen3‑8B on HybridDialogue, while NEIP sees the greatest improvement in comparison between Llama-3.1-8B-Instruct and \textsc{Fine-Refine} on OpendialKG. Generally, \textsc{Fine-Refine} outperforms the RAG baselines G-Retriever and Self-RAG in terms of factual accuracy.

Regarding quality, we note small declines in maintains context and natural (e.g., natural: 2.948 to 2.775, -5.87\% on HybridDialogue, Mistral-7B-Instruct-v0.3) following \textsc{Fine-Refine}. Given that factual errors in knowledge-intensive dialogue can mislead users and are typically judged more harshly than slight reductions in quality, we consider this trade-off acceptable. Moreover, compared to \textsc{Self-Refine}, \textsc{Fine-Refine} yields higher factuality with similar dialogue quality.

Overall, \textsc{Fine-Refine} steadily improves dialogue factuality, increasing the dialogue fact score and decreasing the NEIP, whereas \textsc{Self-Refine} improves both the dialogue fact score and the NEIP. Our framework achieves the best performance among baselines with a trade-off in dialogue quality.

\subsection{Iterative Refinement Results}
We conduct a study on the number of iterations to observe the improvement at each time, shown in Table~\ref{table:iterative_models_llm}. 

The results of two iterative refinement methods with Qwen3-8B are shown in Table~\ref{table:iterative_models_llm}. Our \textsc{Fine-Refine} consistently outperforms \textsc{Self-Refine} on Qwen3-8B in dialogue fact score and NEIP, demonstrating the validity of our proposed framework for factuality. Additionally, we find that improvements in factuality are associated with decreased dialogue qualities, indicating their contrast, and the factors contributing to the results will be analysed in the ablation study below.%\todo{"in the ablation study below" maybe clearer}

In conclusion, iterative refinement leads to a consistent increase in dialogue factuality and sees a slight decrease in dialogue quality.

\begin{table*}[!ht]
\centering
\caption{Ablation on a single component of feedback for HybriDialogue and OpendialKG with LLM-based evaluation.}
\resizebox{0.85\textwidth}{!}{
\begin{tabular}{l|l|cccc|cccc}
\toprule
\multirow{2}{*}{\textbf{\textbf{Model}}} & \multirow{2}{*}{\textbf{Setting}} & \multicolumn{4}{c|}{\textbf{HybriDialogue}} & \multicolumn{4}{c}{\textbf{OpendialKG}} \\
\cmidrule(lr){3-6} \cmidrule(lr){7-10}
 & & \textbf{Fact} & \textbf{NEIP$\downarrow$} & \textbf{Maintains Ctx.} & \textbf{Natural}
   & \textbf{Fact} & \textbf{NEIP$\downarrow$} & \textbf{Maintains Ctx.} & \textbf{Natural} \\
\toprule
\multirow{4}{*}{Mistral-7B-Instruct-v0.3} 
 & \textsc{Fine-Refine}\textsuperscript{1} & 78.99 & \textbf{46.78} & 2.901 & 2.876 & 86.31 & \textbf{48.22} & 2.969 & 2.735 \\
 & Only Response              & 77.87 & 49.99 & \textbf{2.928} & \textbf{2.919} & 85.77 & 51.13 & \textbf{2.987} & \textbf{2.804} \\
 & Only Fact                  & \textbf{79.08} & 47.38 & 2.904 & 2.844 & \textbf{86.90} & 48.30 & 2.973 & 2.677 \\
 & Only Fluency               & 76.75 & 48.19 & 2.918 & 2.890 & 86.77 & 49.65 & 2.971 & 2.797 \\
\midrule

\multirow{4}{*}{Llama-3.1-8B-Instruct} 
 & \textsc{Fine-Refine}\textsuperscript{1} & 83.21 & 44.94 & 2.754 & 2.605 & 87.53 & 40.85 & 2.891 & 2.614 \\
 & Only Response              & 81.28 & 47.37 & 2.778 & \textbf{2.687} & 85.33 & 43.42 & \textbf{2.909} & \textbf{2.694} \\
 & Only Fact                  & \textbf{83.40} & \textbf{44.85} & 2.735 & 2.601 & \textbf{87.76} & \textbf{40.59} & 2.883 & 2.602 \\
 & Only Fluency               & 81.78 & 45.79 & \textbf{2.784} & 2.627 & 86.49 & 42.71 & 2.894 & 2.636 \\
\midrule

\multirow{4}{*}{Qwen3-8B} 
 & \textsc{Fine-Refine}\textsuperscript{1} & 79.39 & 50.08 & 2.842 & 2.881 & \textbf{87.36} & 57.62 & 2.970 & 2.961 \\
 & Only Response              & 74.36 & 53.81 & \textbf{2.894} & \textbf{2.946} & 84.40 & 59.07 & \textbf{2.976} & \textbf{2.970} \\
 & Only Fact                  & \textbf{79.42} & \textbf{49.01} & 2.847 & 2.892 & 86.84 & \textbf{56.81} & 2.975 & 2.962 \\
 & Only Fluency               & 74.34 & 53.96 & 2.862 & 2.915 & 84.62 & 59.04 & 2.975 & 2.965 \\
\bottomrule
\end{tabular}
}
\label{table:ablation_single_weight_llm_full}
\end{table*}

\subsection{Ablation Study}
We conduct an ablation study to analyse the individual components' contributions, as shown in Table~\ref{table:ablation_single_weight_llm_full}. The ablation is conducted with the first refinement iteration.

Using Only Response feedback (without atomic units decomposition) generally yields better dialogue quality (Maintains Ctx. and Natural) but provides smaller factuality gains compared with variants that incorporate factual verification. In contrast, Only Fact achieves the strongest or comparable factuality (Fact / NEIP) but can hurt dialogue quality, highlighting a trade-off between factual correction and conversational naturalness. Finally, Only Fluency does not consistently improve factuality, suggesting that fluency-oriented feedback alone is insufficient for hallucination mitigation. 

Overall, \textsc{Fine-Refine} balances factuality and dialogue quality by combining factuality and fluency signals at the atomic-unit level.

\begin{table}[h!]
\centering
\caption{A case study of dialogue evaluation and refinement collected from Qwen3-8B and corresponding \textsc{Fine-Refine}.}
\resizebox{0.49\textwidth}{!}{
\begin{tabular}{p{1.4cm}p{7.8cm}}
\hline
\textbf{Type} & \textbf{Content} \\
\hline
Dialogue History & Speaker A: What is the decan all about? \newline
Speaker B: The decan are 36 groups of stars (small constellations) used in the Ancient Egyptian astronomy. \newline
Speaker A: How many signs are there? \\
\hline
Original Response & Speaker B: There are 36 signs in the decan, each representing a group of stars. \newline
(\textbf{Dialogue fact score}: 0.5) \newline 
\textbf{Maintains Context}: 2.99, \textbf{natural}: 2.56)\\
\hline
Atomic Unit 1 & Speaker B says there are 36 signs in the decan. \newline
(\textbf{Fact}: False, \textbf{Fluency score}: 0)
 \\
\hline
Atomic Unit 2 & Speaker B says each sign represents a group of stars. \newline
(\textbf{Fact}: True, \textbf{Fluency score}: 1) \\
\hline
Refined Response &
Speaker B: There are 36 decans, each representing a group of stars. \newline
(\textbf{Dialogue fact score}: 1) \newline (\textbf{Maintains Context}: 2.04, \textbf{natural}: 2.02)\\
\hline
\end{tabular}
}
\label{tab:dialogue_results}
\end{table}

\subsection{Case Study}
Table~\ref{tab:dialogue_results} presents a dialogue about \emph{signs}, illustrating the limitation of our current refinement approach, which slightly reduces dialogue quality. 

The original response is generated by Qwen3-8B and contains factual errors. Follow the framework, we decompose it into atomic units, evaluate each for factuality and normalised fluency scores, and feed the scores back for refinement; this raises factuality but slightly reduces naturalness, as reflected in the natural scores. However, we can see that the drop in naturalness is caused by removing the hallucination part, indicating a conflict between factuality and dialogue quality. 

\section{Conclusion}
In scenarios aimed at enhancing factual correctness, RAG methods lack post-verification, and existing refinement approaches focus only on entire responses, ignoring that a single response may contain both correct and incorrect facts. Inspired by the subgoaling strategy \citep{simon1971human}, we address this gap by proposing a fine-grained, iterative framework to refine dialogue responses. Our framework decomposes the response into atomic units, assesses each atomic unit, and provides fine-grained feedback to guide refinement. Results on HybriDialogue and OpendialKG show that our method improves factual accuracy in various LLMs, with the highest gain reaching 7.63 points on the dialogue fact score. This improvement does come with some trade-off in dialogue quality, but results indicate that hallucination mitigation is effective. In future work, we aim to improve both factuality and overall dialogue quality through advanced strategies, such as exploring LLM architectures and better context modelling.

\printbibliography

\end{document}